\documentclass[twoside,11pt]{article}

\usepackage{jmlr2e}
\usepackage[symbol]{footmisc}

\usepackage{tikz}
\usetikzlibrary{shapes,decorations,arrows,calc,arrows.meta,fit,positioning}
\tikzset{
    -Latex,auto,node distance =1 cm and 1 cm,semithick,
    state/.style ={ellipse, draw, minimum width = 0.7 cm},
    point/.style = {circle, draw, inner sep=0.04cm,fill,node contents={}},
    bidirected/.style={Latex-Latex,dashed},
    el/.style = {inner sep=2pt, align=left, sloped}
}

\usepackage{listings}
\usepackage{graphicx} 
\usepackage{multicol}
\usepackage{array}
\usepackage{microtype}
\usepackage{placeins}
\usepackage{scalerel}
\usepackage{booktabs} 
\usepackage{multicol}
\usepackage{scalerel}

\ShortHeadings{Poisoning the Search Space in Neural Architecture Search}{R. Wu, N. Saxena \& R. Jain}
\firstpageno{1}

\begin{document}
\footnote[0]{$^*$ Equal Contribution}
\title{Poisoning the Search Space in Neural Architecture Search}

\author{\name Robert Wu$^{*}$\email rupert.wu@mail.utoronto.ca \\
       \addr Department of Computer Science\\
       University of Toronto\\
       \AND
       \name Nayan Saxena$^{*}$ \email nayan.saxena@mail.utoronto.ca\\
       \addr Department of Statistical Sciences\\
       University of Toronto\\
     \AND
       \name Rohan Jain$^{*}$ \email rohan.jain@mail.utoronto.ca\\
       \addr Departments of Computer Science
       \& Mathematics\\
       University of Toronto\\}

\maketitle

\begin{abstract}
Deep learning has proven to be a highly effective problem-solving tool for object detection and image segmentation across various domains such as healthcare and autonomous driving.  At the heart of this performance lies neural architecture design which relies heavily on domain knowledge and prior experience on the researchers' behalf. More recently, this process of finding the most optimal architectures, given an initial search space of possible operations, was automated by Neural Architecture Search (NAS). In this paper, we evaluate the robustness of one such algorithm  known as Efficient NAS (ENAS) against data agnostic poisoning attacks on the original search space with carefully designed ineffective operations.  By evaluating algorithm performance on the CIFAR-10 dataset, we empirically demonstrate how our novel search space poisoning (SSP) approach and multiple-instance poisoning attacks exploit design flaws in the ENAS controller to result in inflated prediction error rates for child networks. Our results provide insights into the challenges to surmount in using NAS for more adversarially robust architecture search.
\end{abstract}


\section{Introduction}
\begin{figure}[htbp]
    \centering
        \begin{tikzpicture}
    \node[state,rectangle] (k) at (-2,1) {\tiny Search Space $(\mathcal{A})$};
    \node[state,rectangle] (x) at (1,1) {\tiny Search Strategy};
    \node[state,rectangle] (y) at (4,1) {\tiny Performance Estimation};

    \path (x) edge (y);
     \path (x) edge[bend left= 30] node[above, el] {\tiny Architecture $a\in \mathcal{A}$} (y);
     \path  (y) edge[bend left= 30] node[below, el] {\tiny Performance Metric} (x);
    \path (k) edge (x);
    \node[draw=blue,dotted,fit=(x) (y), inner sep=0.2cm] (machine) {};
    \end{tikzpicture}
    \caption{Overview of the NAS framework}
    \label{nas}
\end{figure}
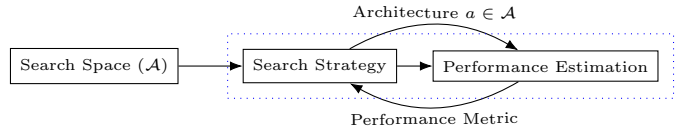

In the modern ecosystem, the problem of finding the most optimal deep learning architectures has been a major focus of the machine learning community.  With applications ranging from speech recognition \citep{hinton2012deep} to image segmentation \citep{krizhevsky2012imagenet}, deep learning has shown the potential to solve pressing issues in several domains including healthcare \citep{wang2016deep,piccialli2021survey} and surveillance \citep{liu2016deep}. However, a major challenge is to find the best architecture design for a given problem. This relies heavily on the researcher's domain knowledge and involves large amounts of trial and error.  More recently, neural architecture search (NAS) algorithms have automated this dynamic process of creating and evaluating new architectures \citep{zoph2016neural,liu2018darts,liu2018progressive}. These algorithms continually sample operations from a predefined search space to construct architectures that best optimize a performance metric over time, eventually converging to the best child architectures.  This intuitive idea, outlined in Figure \ref{nas}, greatly reduces human intervention by restricting human bias in architecture engineering to just the selection of the predefined search space \citep{elsken2019neural}.  

Although NAS has the potential to revolutionize architecture search across industry and research applications, human selection of the search space also presents an open security risk that needs to be evaluated before NAS can be deployed in security-critical domains. Due to the heavy dependence of NAS on the search space, poor search space selection either due to human error or by an adversary has the potential to severely impact the training dynamics of NAS. This can alter or completely reverse the predictive performance of even the most optimal final architectures derived from such a procedure. In this paper, we validate these concerns by evaluating the robustness of one such NAS algorithm known as Efficient NAS (ENAS) \citep{pham2018efficient} against data-agnostic search space poisoning (SSP) attacks.

\section{Related Work}

A comprehensive overview of NAS algorithms can be found in \citet{wistuba2019survey} and \citet{elsken2019neural}, with \citet{chakraborty2018adversarial} summarising advances in adversarial machine learning including poisoning attacks. NAS algorithms have recently been employed in healthcare and applied in various clinical settings for diseases like COVID-19, cancer and cystic fibrosis \citep{van2020automl}. Furthermore, architectures derived from NAS procedures have shown state of the art performance, often outperforming manually created networks in semantic segmentation \citep{chen2018searching}, image classification \citep{real2019regularized, zoph2018learning} and object detection \citep{zoph2018learning}.  With rapid development of emerging NAS methods, recent work by \citet{lindauer2020best} has brought to light some pressing issues pertaining to the lack of rigorous empirical evaluation of existing approaches.
Furthermore, while NAS has been studied to further develop more adversarially robust networks through addition of dense connections \citep{kotyan2019evolving,guo2020meets}, little work has been done in the past to assess the adversarial robustness of NAS itself. Search phase analysis has shown that computationally efficient algorithms such as ENAS are worse at truly ranking child networks due to their reliance on weight sharing \citep{yu2019evaluating}, which can be exploited in an adversarial context. Finally, most traditional poisoning attacks involve injecting mislabeled examples in the training data and have been executed against feature selection methods \citep{xiao2015feature}, support vector machines \citep{biggio2012poisoning} and neural networks \citep{yang2017generative}. To the authors’ knowledge, no study, has approached poisoning in a data-agnostic manner, especially one that involves poisoning the search space in NAS.  In summary, our main contributions through this paper are that:

\begin{itemize}
    \item We emphasise the conceptual significance of designing adversarial poisoning attacks that leverage the original search space and controller design in ENAS.
    \item We propose and develop the theory behind a novel data-agnostic poisoning technique called search space poisoning (SSP) alongside multiple-instance poisoning attacks, as described in Section \ref{ssp}.
    \item  Through our experiments on the CIFAR-10 dataset in Section \ref{experiments} we demonstrate how SSP results in child networks with inflated prediction error rates (up to $\sim80\%$). 
\end{itemize}

\section{Background}
\subsection{Efficient Neural Architecture Search (ENAS)}

\subsubsection{Search Space}
Consider the set $\mathcal{A}$ containing all possible neural network architectures or child models that can be generated. The ENAS search space is then represented as a directed acyclic graph (DAG) denoted by $\mathcal{G}$ which is the superposition of all child models in $\mathcal{A}$. 

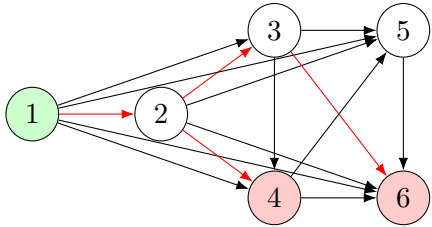
\begin{figure}[htbp]
    \centering
    \resizebox{!}{!}{
    \begin{tikzpicture}
    \node[state,fill=green!20] (1) { $1$ };
    \node[state] (2) [right =of 1] {$2$};
    \node[state] (3) [above right  =of 2,xshift=-0.0cm,yshift=-0.4cm] {$3$};
     \node[state,fill=red!20] (4) [below right =of 2,xshift=-0.0cm,yshift=+0.4cm] {$4$};
    \node[state] (5) [right  =of 3,xshift=-0.0cm,yshift=0cm] {$5$};

    \node[state,fill=red!20] (6) [right  =of 4,xshift=-0.0cm,yshift=0cm] {$6$};

    \path[red] (1) edge node[above] {} (2);
    \path[red] (2) edge node[above] {} (3);
    \path (1) edge node[el,above] {} (4);
    \path (1) edge node[el,above] {} (3);
    \path (1) edge node[el,above] {} (5);
    \path (1) edge node[el,above] {} (6);
    \path[red] (2) edge node[el,above] {} (4);
    \path (2) edge node[el,above] {} (5);
    \path (2) edge node[el,above] {} (6);
    
    \path (3) edge node[el,above] {} (4);
     \path (3) edge node[el,above] {} (5);
      \path[red] (3) edge node[el,above] {} (6);
       \path (4) edge node[el,above] {} (5);
        \path (4) edge node[el,above] {} (6);
        \path (5) edge node[el,above] {} (6);
\end{tikzpicture}}
    \caption{ENAS search space represented as a DAG. Red arrows represent one child model with input node 1 and outputs 4, 6 respectively.}
    \label{fig:dag}
\end{figure}

Every node in Figure 2 represents local computations each having its own parameters with edges representing the flow of information between nodes. Sampled architectures are sub-graphs of $\mathcal{G}$ with parameters being shared amongst child models. Throughout this paper, we focus on the highly effective  original ENAS search space as outlined in \citet{pham2018efficient} denoted by $\hat{\mathcal{S}}$ = \{Identity, 3x3 Separable Convolution, 5x5 Separable Convolution, Max Pooling (3x3), Average Pooling (3x3)\}.              
\subsubsection{Search Strategy}
The ENAS controller is a predefined long short term memory (LSTM) cell which autoregressively samples decisions through softmax classifiers. The central goal of the controller is to search for optimal architectures by generating a child model $a \in \mathcal{G}$, feeding every decision on the previous step as an input embedding into the next step. Our main search strategy throughout this paper will be macro search where  the controller makes two sampling decisions for every layer in the child network: (i) connections to previous nodes for skip connections, and (ii) operations to use from the search space. The model is finally evaluated for its performance which is further used to optimize reward as described next.

\subsubsection{Performance Estimation}  As outlined in \citet{pham2018efficient},
ENAS alternates between training the shared parameters $\omega$ of the child model $\mathbf{m}$ using stochastic gradient descent (SGD), and parameters $\theta$ of the LSTM controller using reinforcement learning (RL). First, keeping $\omega$ fixed, $\theta$ is  trained  with {\small{REINFORCE}} \citep{williams1992simple} and Adam optimiser \citep{kingma2014adam} to maximize the expected reward $\mathbb{E}_{\textbf{m} \sim \pi(\textbf{m}; \theta)}[\mathcal{R}(\textbf{m}, \omega)]$ (validation accuracy); and second, keeping the controller's policy $\pi(\textbf{m}, \theta)$ fixed,  $\omega$ is updated with SGD to minimize expected cross-entropy loss $\mathbb{E}_{\textbf{m} \sim \pi}[\mathcal{L}(\textbf{m}; \omega)]$. It should be noted that different operations associated with the same node in $\mathcal{G}$ have their own unique parameters.

\subsection{Training Data Poisoning}
Traditionally,  training data poisoning is defined as the adversarial contamination of the training set $T\subset \mathcal{D}$ by addition of an extraneous data point $(\textbf{x}_p,\textbf{y}_p)$ which maximizes prediction error across training and validation sets, while significantly impacting loss minimization during training \citep{xiao2015feature, biggio2012poisoning, munoz2017towards, yang2017generative}.  It is assumed here that the data is generated according to an underlying process $f: X \mapsto Y$, given a set $\mathcal{D} = \{\textbf{x}_i, \textbf{y}_i\}_{i=1}^n$ of \textit{i.i.d} samples drawn from $p(X,Y)$, where $X$  and $Y$ are sets containing feature vectors and corresponding target labels respectively. While highly effective, existing poisoning techniques are highly data dependent and operate under the assumption that the attacker has access to training data. A more relaxed assumption would be to decouple the attack modality from training data and make it data agnostic, which is explored in the subsequent section. 

\section{Search Space Poisoning (SSP)}
\label{ssp}
\subsection{General Framework}

\begin{figure}[htbp]
    \centering
        \begin{tikzpicture}
    \node[state,rectangle ,fill=red!20] (j) at (-2,2) {\tiny Poisoning Set $(\mathcal{P})$};
    \node[state,rectangle] (k) at (-2,1) {\tiny Search Space $(\hat{\mathcal{S}})$};
    \node[state,rectangle] (x) at (1,1) {\tiny Search Strategy};
    \node[state,rectangle] (y) at (4,1) {\tiny Performance Estimation};
    \path (x) edge (y);
     \path (x) edge[bend left= 30] node[above, el] {\tiny Architecture} (y);
     \path (y) edge[bend left= 30] node[below, el] {\tiny Performance Metric }  (x);
    \path (k) edge   node[above, el] { ${\scriptscriptstyle  \hat{{S}} \cup {P}}$ }  (x) ;
     \path (j) edge  (k) ;
    \node[draw=blue,dotted,fit=(x) (y), inner sep=0.2cm] (machine) {};
    \end{tikzpicture}
    \caption{Overview of Search Space Poisoning (SSP)}
    \label{nas2}
\end{figure}
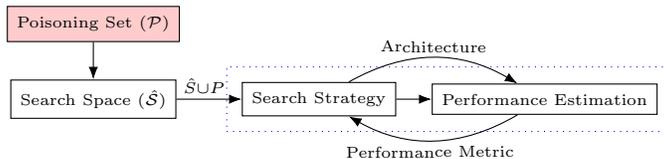

Motivated by the previously described notion of training data poisoning, we introduce search space poisoning (SSP) focused on contaminating the original ENAS search space.  The core idea behind SSP is to inject precisely designed ineffective operations into the ENAS search space to maximize the frequency of poor architectures appearing during training. Our approach exploits the core functionality of the
ENAS controller to sample child networks from a large computational graph of operations by introducing highly ineffective local operations into the search space. On the attacker's behalf, this requires no \textit{a priori} knowledge of the problem domain or training dataset being used, making this new approach more favourable than traditional poisoning attacks. Formally, we describe a poisoned search space as $\mathcal S := \hat{\mathcal{S}} \cup \mathcal P$, where $ \hat{\mathcal{S}}$ denotes the original ENAS search space operations and $\mathcal P$ denotes a non-empty set of poisonings where each poisoning is an ineffective operation. An overview of the SSP approach can be observed Figure \ref{nas2}.

\subsection{Multiple-Instance Poisoning Attacks}
Over the course of training, the LSTM controller paired with the RL search strategy in ENAS develops the ability to sample architectures with operations that most optimally reduce the validation error. As a result, single-instance poisoning attacks might not be as effective due to the tendency of the ENAS controller to draw fewer child networks with the single sub-optimal operation $o_{\scaleto{\mathcal{P}}{3.4pt}} \in \mathcal P$ over time. This behaviour of ENAS results in the algorithm almost entirely discarding networks with the singular $o_{\scaleto{\mathcal{P}}{3.4pt}}$ as training progresses. To circumvent this issue, we propose multiple-instance poisoning which essentially increases the likelihood of  $o_{\scaleto{\mathcal{P}}{3.4pt}}$ being sampled from each poisoned search space. This is achieved by increasing the frequency of sampling $o_{\scaleto{\mathcal{P}}{3.4pt}}$ from $\mathcal S$ through inclusion of multiple-instances of each $o_{\scaleto{\mathcal{P}}{3.4pt}}$ from the poisoning multiset, so-called to allow for duplicate elements. An instance factor $q \in \mathbb N^{> 0}$ would represent instance multiplication of $o_{\scaleto{\mathcal{P}}{3.4pt}}$ in the multiset $q$ times. Henceforth, the probabilities of sampling $o_{\scaleto{\hat\mathcal{{S}}}{5pt}} \in \hat\mathcal{{S}}$ and $o_{\scaleto{\mathcal{P}}{3.4pt}} \in \mathcal{P}$, respectively, are,
\begin{equation}
    \label{prob}
    {Pr}[o_{\scaleto{\hat \mathcal{S}}{5pt}}] := \frac{1}{|\mathcal S| + q|\mathcal P|} < {Pr}[o_{\scaleto{\mathcal{P}}{3.4pt}}] := \frac{q}{|\mathcal S| + q| \mathcal P|} 
\end{equation}
    


From result \ref{prob} it is evident that under a multiple-instance poisoning framework, the probability of sampling poisoned operations is higher as compared to operations in $\hat \mathcal{S}$. Another challenge to overcome is to craft each $o_{\scaleto{\mathcal{P}}{3.4pt}} \in \mathcal P$ such that it counteracts the efficacy of the original operations $o_{\scaleto{\hat \mathcal{S}}{5pt}} \in \hat \mathcal{S}$, which we tackle in the next section.

\subsection{Crafting Poisoning Sets with Operations}

\subsubsection{Identity Operation}
The simplest way to attack the functionality of ENAS is to inject non-operations within the original search space which keep the input and outputs intact. As a result, the controller will sample child models with layers representing computations which preserve the inputs, making the operation highly ineffective within a network architecture.  This goal is fulfilled by the identity operation which has no numerical effect on the inputs with a minimal computational cost. It should also be noted that, the identity operation is not a skip connection. Therefore, we define our first set of poisonings as $ \mathcal P_1 :=$ \{Identity\}.

\subsubsection{Transposed Convolutions} As described earlier, amongst other useful operations the original ENAS search space $\hat{\mathcal{S}}$ also contains 3x3 and 5x5 convolutional layers (separable). With these settings under consideration, a more practical way of poisoning the search space is to reverse the effect of each of these convolutions. Given a normal convolutional layer $g$ and a transposed convolutional layer $h$ with the same parameters except for output channel sizes, $g$ and $h$ are approximate inverses. We achieve our goal of countering the effect of existing convolutions by including transposed convolutions in the set of poisonings resulting in our second poisoning set being  $\mathcal{P}_2 :=$ \{3x3 transposed convolution, 5x5 transposed convolution\}.

\subsubsection{Dropout Layer} While dropout layers have historically been shown to be useful in preventing neural networks from over-fitting \citep{srivastava2014dropout}, a high dropout rate can result in severe information loss leading to poor performance of the overall network. This is because given a dropout probability $p \in \mathbb [0, 1]$, dropout randomly zeroes out some values from the input to decorrelate neurons during training. We hypothesise that including such layers with high dropout probability, such as $p = 0.9$, has the potential to contaminate the search space with irreversible effects on the training dynamics of ENAS. Therefore, our final poisoning set is simply $\mathcal P_3 :=$ \{Dropout ($p=0.9$)\}.


\section{Experiments}

\label{experiments}
\begin{table}[htbp] 
\label{tbl}
  \centering
  \footnotesize
  \resizebox{\textwidth}{!}{
  \begin{tabular}{c|c|c|c|c|c}
    \toprule
{\textsc{Poisoning Set}} &   \textsc{Search Space} & \textsc{Experiment} & \textsc{Poisoning Multiset} & \textsc{Validation Error} & \textsc{Test Error} \\ $\mathcal{P}_i$  & $\mathcal{S}_i$  && $q(\mathcal{P}_i)$ && \\
\hline\hline
  $\phi$ & $\hat{\mathcal{S}}$ & Original  & $\varnothing$ & $19.53$ & $25.33$ \\ 
    \hline
    && 1a & $6(\mathcal P_1)$ & $22.32$ & $27.28$ \\
  $\mathcal P_1 =$ \{{Identity}\}  & $\mathcal{S}_1 = \hat{\mathcal{S}} \cup\mathcal P_1 $& 1b & $36(\mathcal P_1)$  & $37.12$ & $40.87$ \\
    && 1c & $120(\mathcal P_1)$ & $58.67$ & $55.29$ \\
    && 1d & $300(\mathcal P_1)$& $\mathbf{72.60}$ & $\mathbf{69.19}$ \\
    \hline
  && 2a & $1(\mathcal P_2)$ & $20.95$ & $24.25$\\
  $\mathcal{P}_2 =$ \{3x3 transposed convolution, & $\mathcal{S}_2 =\hat{\mathcal{S}} \cup\mathcal P_2$&2b & $6(\mathcal P_2)$ & $32.33$ & $34.78$ \\
  \hspace{3em} 5x5 transposed convolution\}   &&2c  & $20(\mathcal P_2)$  & $33.99$ & $37.05$ \\ 
     &&2d & $50(\mathcal P_2)$ & $\mathbf{68.63}$ & $\mathbf{65.41}$ \\
    \hline
     &&3a & $6(\mathcal P_3)$ & $27.94$ & $34.68$ \\ 
  $\mathcal{P}_3 =$ \{Dropout ($p=0.9$)\} & $\mathcal{S}_3 =\hat{\mathcal{S}} \cup\mathcal P_3$& 3b & $36(\mathcal P_3)$  & $48.17$ & $50.61$  \\
     &&3c  & $120(\mathcal P_3)$ & $73.63$ & $73.70$\\ 
     && 3d & $300(\mathcal P_3)$ & $\mathbf{83.69}$ & $\mathbf{82.07}$ \\
    \hline
     && 4a & $1(\mathcal P_4)$& $25.60$ & $31.81$\\
  $\mathcal P_4 := \mathcal P_1 \cup \mathcal P_2 \cup \mathcal P_3$ & $\mathcal{S}_4 = \hat{\mathcal{S}} \cup\mathcal P_4$& 4b & $6(\mathcal P_4)$ & $36.80$ & $41.81$\\
      && 4c & $20(\mathcal P_4)$ & $\mathbf{69.05}$ & $\mathbf{65.55}$ \\ 
     && 4d  & $50(\mathcal P_4)$ & $\mathbf{68.35}$ & $\mathbf{64.64}$\\ 
    \bottomrule
  \end{tabular}}
    \caption{Summary of experimental search spaces with corresponding final validation and test accuracies.}
\end{table}

\begin{figure*}[htbp]
\centering
\includegraphics[width= 0.7\textwidth]{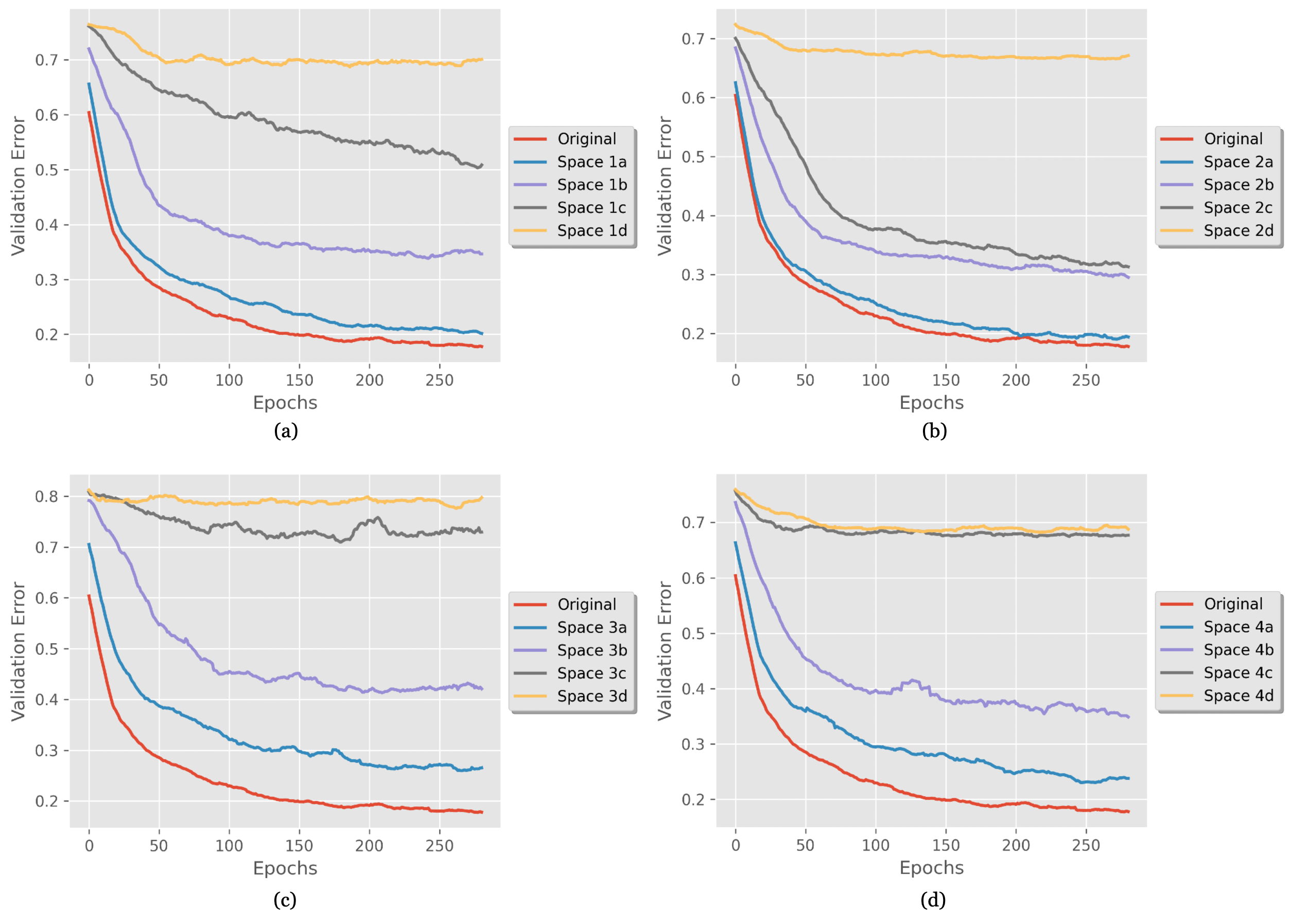}
\caption{ Experimental results for each search space outlined in Table 1 representing moving average of the validation error per 20 epochs for 300 total epochs.}
\label{graphs1}
\end{figure*}

To test the effectiveness of our proposed approach, we designed experiments based on previously described methods as outlined in Table 1. Each experiment involved training ENAS on the CIFAR-10 dataset for 300 epochs on a cluster equipped with an Intel Xeon E5-2620 and Nvidia TITAN Xp GPU (hyperparameters used can be found in Appendix A). {{Code used to run our experiments can be found \href{https://github.com/rusbridger/ENAS-Experiments}{\color{blue}here}.}} Across our experiments, errors increased consistently in relation to the incremental addition of ineffective operations as seen in Table 1 and visualized in Figure \ref{graphs1}(a)-(d). 

\begin{figure*}[htbp]
\centering
\includegraphics[width= 0.7\textwidth]{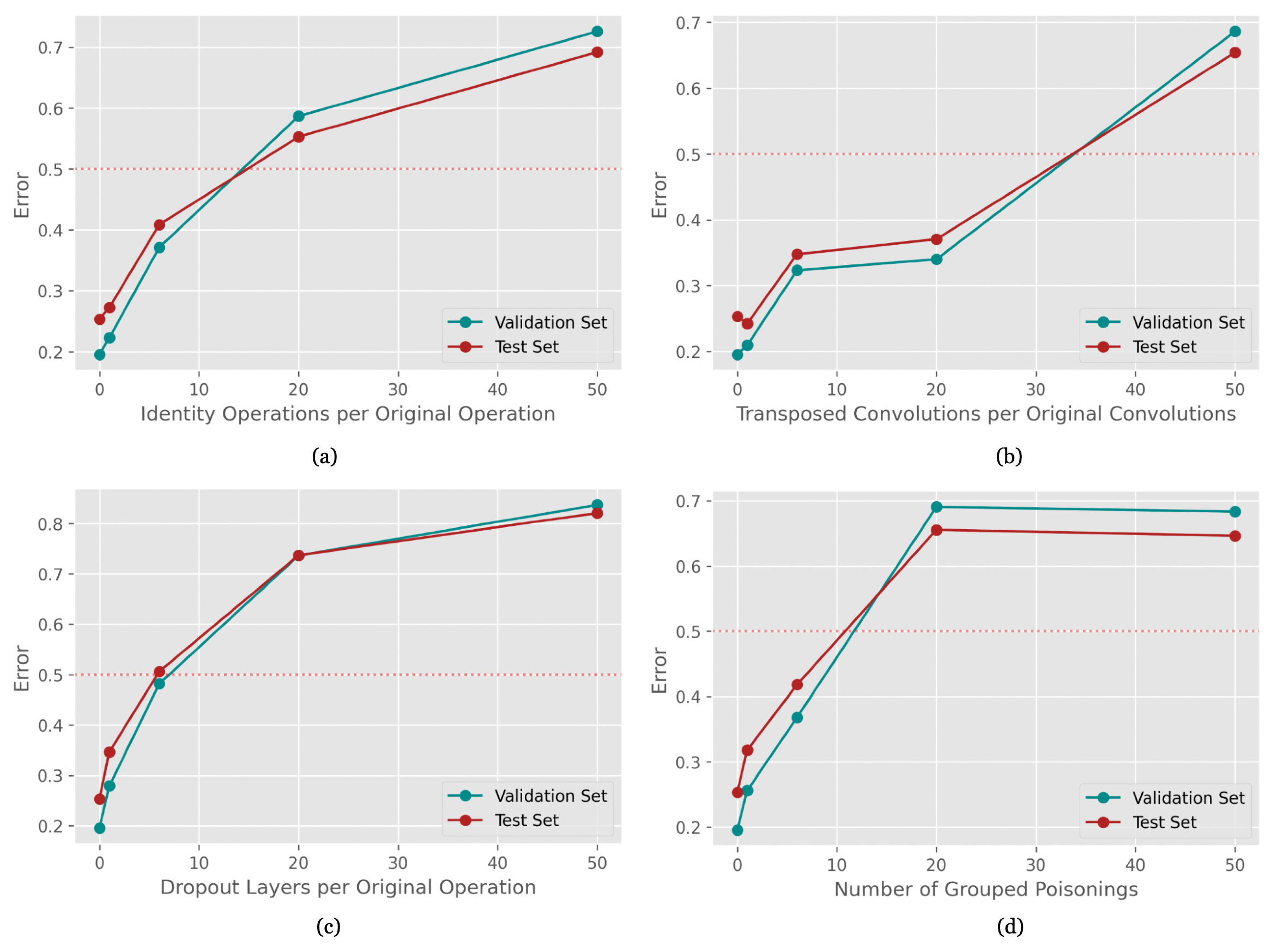}
\caption{Experimental results for each search space outlined in Table 1 representing final validation and test classification errors as a function of multiple operation instances. }
\label{graphs2}
\end{figure*}


\subsection{Identity Operation} Figure \ref{graphs1}(a) shows that instance-multiplied identity operations increase the error considerably. Experiments 1b, 1c, 1d have several identity operations and resulted in high errors, with the extreme $69.19\%$ in experiment 1d. In contrast, experiment 1a only has one identity per original operation and only raised error slightly to $27.28\%$. These results reinforce our hypothesis laid in equation \ref{prob}. Figure \ref{graphs2}(a) shows that excessive poisonings will result in diminishing returns.

\subsection{Transposed Convolutions} Instance-multiplying transposed convolutions had mostly similar results of progressively increasing error as seen in Figure \ref{graphs1}(b). We note that an instance factor of $50$ (experiment 2d) results in an extreme increase in error at $68.83\%$; similar behaviour was observed in our other experiments but to a lesser degree. Figure \ref{graphs2}(b) further shows that between the first four experiments, the increase in error slows down. However, the $100$ transposed convolutions in experiment 2d show a staggering $28.36\%$ increase in error. This inflection presents a desirable point for attacks to exploit.

\subsection{Dropout Layer} Instance-multiplying dropout operations exhibited a similar pattern in validation to the previous operations, but the poisoning seemed to inflate the errors to a greater degree as seen in Figure \ref{graphs1}(c). Figure \ref{graphs2}(c) shows the experiments progressively worsening in error with experiment 3d hitting $83.69\%$ validation and $82.07\%$ test errors. We also observe that adding further dropout, like $300$, results in smaller increases in error, like identity and unlike transposed convolutions operations. Dropout's pattern is similar to identity, but its effect on ENAS is more detrimental.

\subsection{Grouped Operations} Graphing the validation error shows a sharper increase in error, implying that mixing different ineffective operations is more detrimental to ENAS than including several instances of the same operation. In reviewing Figures \ref{graphs1}(d) and \ref{graphs2}(d), we note that the $20$ group poisonings in experiment 4c are about as effective as $300$ identity or $100$ transposed convolution operations (experiments 1d, 2d), and more effective than $36$ dropout operations (experiment 3b). We also observe that experiments 4c, 4d were very close in training and final errors; the final errors were $65.55\%$ and $64.64\%$, respectively. So by factor $20$ in experiment 4c, we have reached the poisoning saturation point. The conclusion we draw is that grouping a variety of poisonings is more efficient in attacking ENAS than over-stacking the same poisoning.

\subsection{Summary} Consistent with the earlier findings in \citet{yu2019evaluating}, our results highlight how the controller's dependence on parameter sharing leads to inaccurate predictions. We successfully demonstrate how using the same weights, although computationally cheap, compromises the functionality of ENAS when injected with poor operations. SSP successfully leverages the inability of ENAS to alternate between weights shared across effective and ineffective operations as demonstrated by experimental results.




\section{Conclusion}

\label{conclusion}
NAS algorithms present an important opportunity for researchers and industry leaders by enabling the automated creation of optimal architectures. Judging by the current pace of research, the future of automated architecture engineering seems bright with many major milestones that will pave the way forward towards more optimal architectures. However, it is also important to evaluate obvious vulnerabilities in these systems which can result in unforeseen model outcomes if not dealt with beforehand. In this paper, we focused on examining the robustness of ENAS under our newly proposed SSP paradigm. We found that infecting the original search space resulted in child architectures that were highly inaccurate in their predictive abilities. Moreover, our carefully designed poisoning sets demonstrated the potential to make it incredibly easy for an attacker with no prior knowledge or access to the training data to still drastically impact the quality of child networks. These findings pave the way for machine learning researchers to explore improvements to the search space and controller design for more adversarially robust search. Finally, our results also present an opportunity for researchers to extend similar ideas to other NAS methods.

\section*{Acknowledgements}
The authors would like to thank George-Alexandru Adam (Vector Institute; University of Toronto) for valuable comments and stimulating discussions that greatly influenced this paper. We are also grateful to Kanav Singla (University of Toronto) and Benjamin Zhuo (University of Toronto) for their initial contributions to the codebase.

\vskip 0.2in
\bibliography{sample}

\begin{thebibliography}{27}
\providecommand{\natexlab}[1]{#1}
\providecommand{\url}[1]{\texttt{#1}}
\expandafter\ifx\csname urlstyle\endcsname\relax
  \providecommand{\doi}[1]{doi: #1}\else
  \providecommand{\doi}{doi: \begingroup \urlstyle{rm}\Url}\fi

\bibitem[Biggio et~al.(2012)Biggio, Nelson, and Laskov]{biggio2012poisoning}
Battista Biggio, Blaine Nelson, and Pavel Laskov.
\newblock Poisoning attacks against support vector machines.
\newblock \emph{arXiv preprint arXiv:1206.6389}, 2012.

\bibitem[Chakraborty et~al.(2018)Chakraborty, Alam, Dey, Chattopadhyay, and
  Mukhopadhyay]{chakraborty2018adversarial}
Anirban Chakraborty, Manaar Alam, Vishal Dey, Anupam Chattopadhyay, and Debdeep
  Mukhopadhyay.
\newblock Adversarial attacks and defences: A survey.
\newblock \emph{arXiv preprint arXiv:1810.00069}, 2018.

\bibitem[Chen et~al.(2018)Chen, Collins, Zhu, Papandreou, Zoph, Schroff, Adam,
  and Shlens]{chen2018searching}
Liang-Chieh Chen, Maxwell~D Collins, Yukun Zhu, George Papandreou, Barret Zoph,
  Florian Schroff, Hartwig Adam, and Jonathon Shlens.
\newblock Searching for efficient multi-scale architectures for dense image
  prediction.
\newblock \emph{arXiv preprint arXiv:1809.04184}, 2018.

\bibitem[Elsken et~al.(2019)Elsken, Metzen, Hutter, et~al.]{elsken2019neural}
Thomas Elsken, Jan~Hendrik Metzen, Frank Hutter, et~al.
\newblock Neural architecture search: A survey.
\newblock \emph{J. Mach. Learn. Res.}, 20\penalty0 (55):\penalty0 1--21, 2019.

\bibitem[Guo et~al.(2020)Guo, Yang, Xu, Liu, and Lin]{guo2020meets}
Minghao Guo, Yuzhe Yang, Rui Xu, Ziwei Liu, and Dahua Lin.
\newblock When nas meets robustness: In search of robust architectures against
  adversarial attacks.
\newblock In \emph{Proceedings of the IEEE/CVF Conference on Computer Vision
  and Pattern Recognition}, pages 631--640, 2020.

\bibitem[Hinton et~al.(2012)Hinton, Deng, Yu, Dahl, Mohamed, Jaitly, Senior,
  Vanhoucke, Nguyen, Sainath, et~al.]{hinton2012deep}
Geoffrey Hinton, Li~Deng, Dong Yu, George~E Dahl, Abdel-rahman Mohamed, Navdeep
  Jaitly, Andrew Senior, Vincent Vanhoucke, Patrick Nguyen, Tara~N Sainath,
  et~al.
\newblock Deep neural networks for acoustic modeling in speech recognition: The
  shared views of four research groups.
\newblock \emph{IEEE Signal processing magazine}, 29\penalty0 (6):\penalty0
  82--97, 2012.

\bibitem[Kingma and Ba(2014)]{kingma2014adam}
Diederik~P Kingma and Jimmy Ba.
\newblock Adam: A method for stochastic optimization.
\newblock \emph{arXiv preprint arXiv:1412.6980}, 2014.

\bibitem[Kotyan and Vargas(2019)]{kotyan2019evolving}
Shashank Kotyan and Danilo~Vasconcellos Vargas.
\newblock Evolving robust neural architectures to defend from adversarial
  attacks.
\newblock \emph{arXiv e-prints}, pages arXiv--1906, 2019.

\bibitem[Krizhevsky et~al.(2012)Krizhevsky, Sutskever, and
  Hinton]{krizhevsky2012imagenet}
Alex Krizhevsky, Ilya Sutskever, and Geoffrey~E Hinton.
\newblock Imagenet classification with deep convolutional neural networks.
\newblock \emph{Advances in neural information processing systems},
  25:\penalty0 1097--1105, 2012.

\bibitem[Lindauer and Hutter(2020)]{lindauer2020best}
Marius Lindauer and Frank Hutter.
\newblock Best practices for scientific research on neural architecture search.
\newblock \emph{Journal of Machine Learning Research}, 21\penalty0
  (243):\penalty0 1--18, 2020.

\bibitem[Liu et~al.(2018{\natexlab{a}})Liu, Zoph, Neumann, Shlens, Hua, Li,
  Fei-Fei, Yuille, Huang, and Murphy]{liu2018progressive}
Chenxi Liu, Barret Zoph, Maxim Neumann, Jonathon Shlens, Wei Hua, Li-Jia Li,
  Li~Fei-Fei, Alan Yuille, Jonathan Huang, and Kevin Murphy.
\newblock Progressive neural architecture search.
\newblock In \emph{Proceedings of the European Conference on Computer Vision
  (ECCV)}, pages 19--34, 2018{\natexlab{a}}.

\bibitem[Liu et~al.(2018{\natexlab{b}})Liu, Simonyan, and Yang]{liu2018darts}
Hanxiao Liu, Karen Simonyan, and Yiming Yang.
\newblock Darts: Differentiable architecture search.
\newblock \emph{arXiv preprint arXiv:1806.09055}, 2018{\natexlab{b}}.

\bibitem[Liu et~al.(2016)Liu, Liu, Mei, and Ma]{liu2016deep}
Xinchen Liu, Wu~Liu, Tao Mei, and Huadong Ma.
\newblock A deep learning-based approach to progressive vehicle
  re-identification for urban surveillance.
\newblock In \emph{European conference on computer vision}, pages 869--884.
  Springer, 2016.

\bibitem[Mu{\~n}oz-Gonz{\'a}lez et~al.(2017)Mu{\~n}oz-Gonz{\'a}lez, Biggio,
  Demontis, Paudice, Wongrassamee, Lupu, and Roli]{munoz2017towards}
Luis Mu{\~n}oz-Gonz{\'a}lez, Battista Biggio, Ambra Demontis, Andrea Paudice,
  Vasin Wongrassamee, Emil~C Lupu, and Fabio Roli.
\newblock Towards poisoning of deep learning algorithms with back-gradient
  optimization.
\newblock In \emph{Proceedings of the 10th ACM Workshop on Artificial
  Intelligence and Security}, pages 27--38, 2017.

\bibitem[Pham et~al.(2018)Pham, Guan, Zoph, Le, and Dean]{pham2018efficient}
Hieu Pham, Melody Guan, Barret Zoph, Quoc Le, and Jeff Dean.
\newblock Efficient neural architecture search via parameters sharing.
\newblock In \emph{International Conference on Machine Learning}, pages
  4095--4104. PMLR, 2018.

\bibitem[Piccialli et~al.(2021)Piccialli, Di~Somma, Giampaolo, Cuomo, and
  Fortino]{piccialli2021survey}
Francesco Piccialli, Vittorio Di~Somma, Fabio Giampaolo, Salvatore Cuomo, and
  Giancarlo Fortino.
\newblock A survey on deep learning in medicine: Why, how and when?
\newblock \emph{Information Fusion}, 66:\penalty0 111--137, 2021.

\bibitem[Real et~al.(2019)Real, Aggarwal, Huang, and Le]{real2019regularized}
Esteban Real, Alok Aggarwal, Yanping Huang, and Quoc~V Le.
\newblock Regularized evolution for image classifier architecture search.
\newblock In \emph{Proceedings of the aaai conference on artificial
  intelligence}, volume~33, pages 4780--4789, 2019.

\bibitem[Srivastava et~al.(2014)Srivastava, Hinton, Krizhevsky, Sutskever, and
  Salakhutdinov]{srivastava2014dropout}
Nitish Srivastava, Geoffrey Hinton, Alex Krizhevsky, Ilya Sutskever, and Ruslan
  Salakhutdinov.
\newblock Dropout: a simple way to prevent neural networks from overfitting.
\newblock \emph{The journal of machine learning research}, 15\penalty0
  (1):\penalty0 1929--1958, 2014.

\bibitem[van~der Schaar(2020)]{van2020automl}
Mihaela van~der Schaar.
\newblock Automl and interpretability: Powering the machine learning revolution
  in healthcare.
\newblock In \emph{Proceedings of the 2020 ACM-IMS on Foundations of Data
  Science Conference}, pages 1--1, 2020.

\bibitem[Wang et~al.(2016)Wang, Khosla, Gargeya, Irshad, and
  Beck]{wang2016deep}
Dayong Wang, Aditya Khosla, Rishab Gargeya, Humayun Irshad, and Andrew~H Beck.
\newblock Deep learning for identifying metastatic breast cancer.
\newblock \emph{arXiv preprint arXiv:1606.05718}, 2016.

\bibitem[Williams(1992)]{williams1992simple}
Ronald~J Williams.
\newblock Simple statistical gradient-following algorithms for connectionist
  reinforcement learning.
\newblock \emph{Machine learning}, 8\penalty0 (3-4):\penalty0 229--256, 1992.

\bibitem[Wistuba et~al.(2019)Wistuba, Rawat, and Pedapati]{wistuba2019survey}
Martin Wistuba, Ambrish Rawat, and Tejaswini Pedapati.
\newblock A survey on neural architecture search.
\newblock \emph{arXiv preprint arXiv:1905.01392}, 2019.

\bibitem[Xiao et~al.(2015)Xiao, Biggio, Brown, Fumera, Eckert, and
  Roli]{xiao2015feature}
Huang Xiao, Battista Biggio, Gavin Brown, Giorgio Fumera, Claudia Eckert, and
  Fabio Roli.
\newblock Is feature selection secure against training data poisoning?
\newblock In \emph{International Conference on Machine Learning}, pages
  1689--1698. PMLR, 2015.

\bibitem[Yang et~al.(2017)Yang, Wu, Li, and Chen]{yang2017generative}
Chaofei Yang, Qing Wu, Hai Li, and Yiran Chen.
\newblock Generative poisoning attack method against neural networks.
\newblock \emph{arXiv preprint arXiv:1703.01340}, 2017.

\bibitem[Yu et~al.(2019)Yu, Sciuto, Jaggi, Musat, and
  Salzmann]{yu2019evaluating}
Kaicheng Yu, Christian Sciuto, Martin Jaggi, Claudiu Musat, and Mathieu
  Salzmann.
\newblock Evaluating the search phase of neural architecture search.
\newblock \emph{arXiv preprint arXiv:1902.08142}, 2019.

\bibitem[Zoph and Le(2016)]{zoph2016neural}
Barret Zoph and Quoc~V Le.
\newblock Neural architecture search with reinforcement learning.
\newblock \emph{arXiv preprint arXiv:1611.01578}, 2016.

\bibitem[Zoph et~al.(2018)Zoph, Vasudevan, Shlens, and Le]{zoph2018learning}
Barret Zoph, Vijay Vasudevan, Jonathon Shlens, and Quoc~V Le.
\newblock Learning transferable architectures for scalable image recognition.
\newblock In \emph{Proceedings of the IEEE conference on computer vision and
  pattern recognition}, pages 8697--8710, 2018.

\end{thebibliography}

\newpage
\appendix
\section*{Appendix}
\setcounter{figure}{0}
\setcounter{table}{0}

\subsection*{A. Hyperparameters} \label{table:hparams}

\begin{table}[htbp]
\caption{Summary of experiment hyperparameters }
  \centering
  \small{
  \begin{tabular}{cc}
    \toprule
    \textsc{Hyperparameter} & \textsc{Value} \\
    \hline
    \midrule
    \texttt{search\_for} & macro \\
    \hline
    \texttt{batch\_size} & $128$ \\
    \texttt{search\_for} & $300$ \\
    \texttt{seed} & $69$ \\
    \texttt{cutout} & $0$ \\
    \texttt{fixed\_arc} & False \\
    \hline
    \texttt{child\_num\_layers} & $12$ \\
    \texttt{child\_out\_filters} & $36$ \\
    \texttt{child\_grad\_bound} & $5.0$ \\
    \texttt{child\_l2\_reg} & $0.00025$ \\
    \texttt{child\_keep\_prob} & $0.9$ \\
    \texttt{child\_lr\_max} & $0.05$ \\
    \texttt{child\_lr\_min} & $0.0005$ \\
    \texttt{child\_lr\_T} & $10$ \\
    \hline
    \texttt{controller\_lstm\_size} & $64$ \\
    \texttt{controller\_lstm\_num\_layers} & $1$ \\
    \texttt{controller\_entropy\_weight} & $0.0001$ \\
    \texttt{controller\_train\_every} & $1$ \\
    \texttt{controller\_num\_aggregate} & $20$ \\
    \texttt{controller\_train\_steps} & $50$ \\
    \texttt{controller\_lr} & $0.001$ \\
    \texttt{controller\_tanh\_constant} & $1.5$ \\
    \texttt{controller\_op\_tanh\_reduce} & $2.5$ \\
    \texttt{controller\_skip\_target} & $0.4$ \\
    \texttt{controller\_skip\_weight} & $0.8$ \\
    \texttt{controller\_bl\_dec} & $0.99$ \\
    \hline
    \texttt{p} (Dropout Rate) & $0.9$ \\
    \bottomrule
  \end{tabular}}
\end{table}

\end{document}